\documentclass{article}

\usepackage[preprint]{corl_2026} 
\usepackage{enumitem}
\usepackage{enumerate}
\usepackage{microtype}
\usepackage{graphicx}
\usepackage[font=small]{subcaption}
\usepackage{booktabs}
\usepackage{url}
\usepackage{hyperref}
\usepackage{titletoc}

\usepackage{amsmath}
\usepackage{amssymb}
\usepackage{bbm}
\usepackage{amsfonts}
\usepackage{mathtools}
\usepackage{amsthm}

\usepackage{multirow}
\usepackage{makecell}
\usepackage{nicefrac}

\usepackage{algorithm}
\usepackage{algpseudocode}
\usepackage{float}

\title{Steering Generative Reinforcement Learning into Stable Robotic Controller}

\author{%
   Yixuan Wang$^{1, *}$\hspace{1em} Shutong Ding$^{1,}$\thanks{Equal contribution. $\dag$Corresponding author.} \hspace{1em} Ke Hu$^{1, *}$ \hspace{1em} Tianxiang Gui$^{1}$ \\ \textbf{Jingya Wang$^{1}$ \hspace{1em} Ye Shi$^{1,\dag}$ } \\ 
  \vspace{1pt}\\
  $^1$ShanghaiTech University \hspace{1em} \\
  \vspace{1pt}\\
  \texttt{ \{wangyx12022, dingsht, huke2024\}@shanghaitech.edu.cn} \\
  \texttt{\{wangjingya, shiye\}@shanghaitech.edu.cn} \\
}

\begin{document}
\maketitle


\begin{abstract}
Diffusion and flow-based generative policies provide a powerful policy class for reinforcement learning by inducing rich stochastic exploration through iterative action generation. However, the stochasticity of diffusion policies is not suitable for stable and precise control in high-dimensional robotic systems, where small action variations can accumulate into inconsistent motion and reduced robustness. To address this issue, we propose SteerGenPO, a latent-space reinforcement learning framework that steers a trained generative policy into a robust deterministic robotic controller. The key idea is to replace stochastic latent sampling of the trained generative policy with a learned latent actor that predicts a state-dependent latent input for the generative policies. This separates exploration and control: stochastic generative sampling provides diverse action proposals during policy learning, while deterministic latent steering provides stable and adaptive control at deployment. We evaluate SteerGenPO on six Isaac Lab benchmarks and a Unitree G1 locomotion task. The results show SteerGenPO improves over both classical RL and generative RL baselines, while its deterministic latent steering produces more stable inference-time behaviors and more reliable command responses.

\end{abstract}



\section{Introduction}

Recent advances in embodied intelligence have shown that reinforcement learning (RL) algorithms can learn effective control policies across humanoid locomotion, robot arm manipulation, and dexterous hand control~\citep{makoviychuk2021isaac,zhuang2024humanoid}. However, most of these methods use Gaussian policies like PPO~\citep{schulman2017proximal}, which parameterize a unimodal action distribution. While this design is simple and convenient, its unimodal structure can be restrictive in high-dimensional robotic control, where the underlying action distribution may be complex and multi-modal rather than concentrated around a single mode. In parallel, diffusion and flow-based RL methods~\citep{ren2024diffusion,ding2025genpo,mcallister2025flow,yi2026flow} have emerged as a more expressive alternative. By generating actions through iterative denoising or flow transformations, these generative policies can model complex, multi-modal action distributions. Compared with classical PPO policies, their multimodality also provides stronger exploration during training by producing diverse action proposals.

Although diffusion-based RL provides expressive policies and strong exploration during training, we find that the multi-modal stochasticity of diffusion can be harmful when the policy is deployed on real-world robots. In contact-rich and high-dimensional systems, small action variations may accumulate into unstable contacts, inconsistent motion, unnecessary torque oscillations, and reduced robustness. One common way to avoid this randomness is to use a fixed latent input, such as the zero vector in deterministic deployment variants like~\citep{yi2026flow}. Although this reduces action variance, it does not intrinsically solve the deployment problem. The zero vector is only the center of the latent prior, not a latent input optimized for task return, and after the nonlinear generative transformation, it may decode to a suboptimal action. As a result, deterministic zero-sampling deployment can improve stability but often sacrifices policy performance.

To address this problem, we propose \emph{SteerGenPO}, a latent-space reinforcement learning framework for steering generative RL policies into stable robotic controllers. SteerGenPO consists of two stages: first, a generative policy is trained with RL, like GenPO~\citep{ding2025genpo}, to provide diverse action proposals and discover high-return behaviors; second, a state-conditioned latent actor is learned to output a latent input for the trained generative policy. In that case, the latent actor replaces both random latent sampling and fixed zero-latent inference at deployment, turning the generative policy into a deterministic stable controller while preserving the expressiveness of the generative policy. To the best of our knowledge, SteerGenPO is the first latent-space RL method designed for reinforcement-learned diffusion or flow policies. Notably, it differs from recent diffusion steering work such as DSRL~\citep{wagenmaker2025steering}, which applies RL in the latent-noise space of a pretrained behavior-cloned diffusion policy. Compared with DSRL, SteerGenPO applies an on-policy paradigm in the second stage and does not require distilling an additional value network for the latent actor. Besides, we also introduce a latent regularization loss that constrains exploration within the reliable support of the latent space, further improving the stability of policy optimization. Finally, we confirm the efficiency of the proposed SteerGenPO in both IsaacLab benchmarks and Unitree G1 locomotion tasks. Our contribution is threefold:

\begin{itemize}[leftmargin=4pt, rightmargin=4pt]
    \item We identify stochastic latent sampling as a key issue for deploying diffusion reinforcement learning policies in real-time robot control, where stable and precise actions are required.
    \item We propose SteerGenPO, a latent-space RL framework that trains a latent-action actor to find optimal deterministic latent noise for generative policies, yielding a more stable and higher-performing controller for robot control.
    \item We conduct extensive evaluations in both Isaac Lab robotic control tasks and a real-world Unitree G1 humanoid locomotion task. The results show that SteerGenPO consistently outperforms classical and diffusion RL baselines, demonstrating its effectiveness for stable robot control.
\end{itemize}
\section{Related Work}
\label{sec:related_work}

\paragraph{Diffusion-Based Reinforcement Learning}

Diffusion models represent complex distributions through iterative denoising~\citep{ho2020denoising,song2020denoising,song2020score}, motivating diffusion policies for action or trajectory modeling in robot learning~\citep{chi2023diffusion,wangdiffusion,janner2022planning}. Much prior work focuses on offline RL, using diffusion models as expressive behavior priors or policy classes~\citep{wangdiffusion,kang2023efficient,maodiffusion}. Beyond offline learning, DIPO and QVPO study online model-free diffusion-policy RL~\citep{yang2023policy,dingdiffusion}, while Q-score matching trains diffusion policies from rewards in an off-policy setting~\citep{psenka2024learning}. GenPO enables diffusion and flow policies to be trained in an on-policy RL loop by handling the likelihood, entropy, and KL terms required by PPO-style updates~\citep{ding2025genpo}. These works show the value of stochastic generative sampling for policy learning, but leave open how such policies should be used when stable and repeatable inference is required.

\paragraph{Latent Space Optimization for Diffusion Models}

Latent and noise-space optimization has been used to control generative models without retraining them. Guidance methods modify denoising or sampling directions in image and conditional generation~\citep{dhariwal2021diffusion,ho2021classifier}, while diffusion inversion maps samples back to noise trajectories for reconstruction and editing~\citep{wallace2023edict}. Similar ideas appear in planning and robot motion generation, where generative priors are used to sample or optimize structured trajectories~\citep{janner2022planning,carvalho2023motion}. For diffusion policies, the latent input determines which action sample emerges. DSRL starts from a behavior-cloned diffusion policy and learns an RL policy over its latent-noise input to improve the pretrained BC policy~\citep{wagenmaker2025steering}, showing that latent-noise optimization can improve diffusion policies without updating the full generative model.

\paragraph{Diffusion Policies for Humanoid Locomotion}
RL has enabled stable humanoid locomotion and whole-body behaviors~\citep{zhuang2024humanoid,radosavovic2024learning,zhang2024wococo,long2024learninghumanoidlocomotionperceptive}, typically using compact policies with standard action distributions. Diffusion-based locomotion provides a more expressive alternative, with applications to real-time legged control, bipedal locomotion on unseen terrain, and legged navigation or motion generation~\citep{huangdiffuseloco,mothish2024birodiff,liu2024dipper,yuan2024preference,tevet2024closd, fan2025one}. These works show the promise of generative models for locomotion, but stochastic sampling can introduce variability undesirable for precise and consistent control.

Overall, prior work shows how to train expressive diffusion or flow policies, optimize latent or noise variables, and apply generative policies to robot control. SteerGenPO addresses a different problem: making an on-policy generative policy stable and deterministic at inference time. Unlike DSRL, which adapts a behavior-cloned diffusion policy through latent-noise RL, SteerGenPO starts from a policy trained by on-policy generative RL and learns a latent selection policy for deterministic inference. By replacing prior-sampled stochastic inference with state-conditioned deterministic latent selection, SteerGenPO preserves generative policy expressivity while improving stability, precision, and repeatability. Our path-following and Unitree G1 experiments further validate these benefits.
\section{Preliminaries}
\label{sec:preliminaries}

\subsection{Generative Diffusion and Flow Policies}

Generative action policies model the action distribution through a latent sampling process rather than a simple unimodal Gaussian. Diffusion policies generate samples by starting from Gaussian noise and applying a learned denoising process~\citep{ho2020denoising,song2020score,chi2023diffusion}. In the conditional policy setting, this process starts from $a_K\sim\mathcal{N}(0,I)$ and iteratively denoises it into an action or action sequence $a_0$, allowing diffusion policies to represent multi-modal action distributions.

Flow-based policies provide a continuous-time alternative that transports a simple prior distribution to the action distribution through a learned vector field~\citep{lipmanflow}. A flow model learns a state-conditioned vector field $v_\theta(x_t,t,s)$ and generates samples by transporting $x_0\sim p_0$ along $dx_t/dt=v_\theta(x_t,t,s)$. Both diffusion and flow policies can therefore be viewed as latent-variable generators,
\begin{equation}
    z\sim p_0(z),
    \qquad
    a=G_\theta(s,z),
    \label{eq:generic_generator}
\end{equation}
where $p_0$ is a simple prior and $G_\theta$ is a state-conditioned generative map. This abstraction describes how generative policies are trained with reinforcement learning and why latent selection remains a separate inference-time problem.

\subsection{Generative Policies for On-Policy Reinforcement Learning}

The unified generator form in Eq.~\eqref{eq:generic_generator} provides a common view of diffusion and flow policies. To use such policies in on-policy RL, however, one must compute the action density induced by the latent sampling process. The generator induces an implicit action distribution by pushing the latent prior through the state-conditioned map. On-policy likelihood-ratio methods such as PPO~\citep{schulman2017proximal} require action log-likelihoods, entropy estimates, and policy ratios. These quantities are straightforward for Gaussian policies but nontrivial for iterative generative policies.

Generative policy optimization methods address this difficulty by constructing invertible generative mappings or by using exact inversion, so that generated actions can be related back to their latent variables~\citep{dinh2016density,ding2025genpo}. If $G_\theta(s,\cdot)$ is invertible for a fixed state $s$, the corresponding latent can be recovered as $z=G_\theta^{-1}(s,a)$, and the action likelihood can be computed by change of variables:
\begin{equation}
    \log \pi_\theta(a\mid s)
    =
    \log p_0\!\left(G_\theta^{-1}(s,a)\right)
    +
    \log
    \left|
    \det
    \frac{\partial G_\theta^{-1}(s,a)}
         {\partial a}
    \right|.
    \label{eq:generic_change_of_variables}
\end{equation}
This makes likelihood-related terms tractable for on-policy updates while retaining the expressive action distributions of generative policies.

Under this view, the generative policy is trained to improve the expected value of actions obtained by sampling latents from the prior:
\begin{equation}
    \max_\theta\;
    \mathbb{E}_{s\sim d^{\pi_\theta},\,z\sim p_0}
    \left[
        Q^{\pi_\theta}(s,G_\theta(s,z))
    \right].
    \label{eq:prior_sampled_objective}
\end{equation}
Thus, stochastic latent sampling is useful during policy learning. However, this objective optimizes the generator under prior-sampled latents and does not specify which latent input should be used for deterministic inference. This motivates the latent-selection problem addressed by SteerGenPO, where we denote the trained executable-action decoder as $D_\theta(s,z)$.

\section{Method}
\label{sec:method}

\begin{figure}
    \centering
    \includegraphics[width=\linewidth]{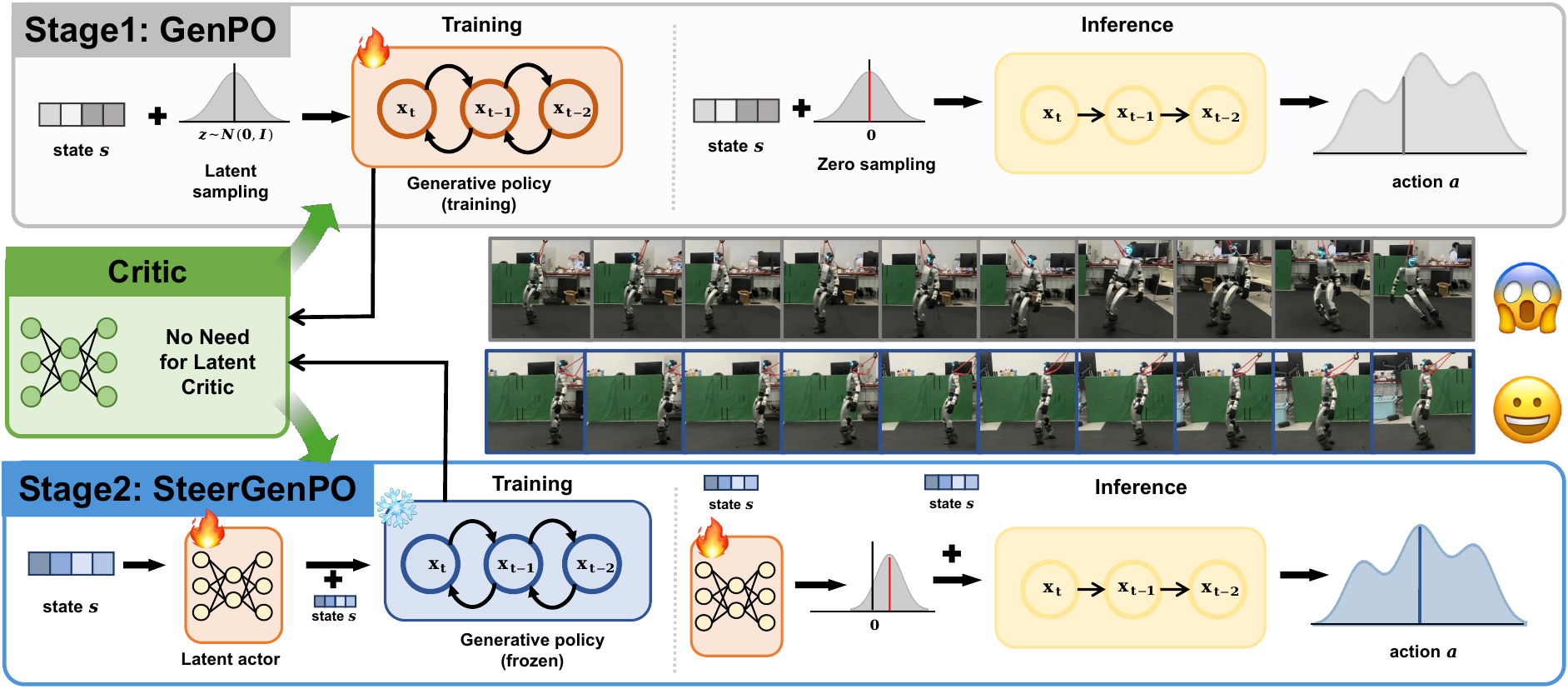}
    \caption{Overview of the proposed latent steering framework, where stochastic diffusion sampling is used for exploration during reinforcement learning, while a state-conditioned latent actor replaces random latent sampling at deployment to produce stable and reproducible humanoid locomotion actions.} 
    \label{fig:placeholder}
\end{figure}

We propose SteerGenPO as two reinforcement learning problems defined over different optimization spaces. The first stage follows GenPO and performs policy optimization in the action space of a
flow-based generative policy. This stage learns an expressive stochastic action generator by updating the policy parameters with on-policy rollouts. The second stage freezes this generator and performs
policy optimization in its latent input space. The goal is no longer to reshape the entire action distribution, but to learn which latent input should be selected for each state so that the frozen generator produces higher-value actions.
\subsection{The Latent Selection Problem in GenPO}

The first stage follows GenPO and trains a stochastic generative policy with on-policy rollouts. Let $\mathcal M=(\mathcal S,\mathcal A,P,r,\gamma)$ denote the original control problem, where
$s\in\mathcal S$ is the state and $a\in\mathcal A\subset\mathbb R^d$ is the executable action. 
Given a state $s$, the policy samples a latent variable from a fixed Gaussian prior, $ z \sim p_0(z)=\mathcal N(0,I)$, and maps it to an executable action $\tilde a = F_\theta(s,z)=(x,y)$, $a=h(\tilde a)=\frac{x+y}{2}$.
The GenPO objective can be viewed abstractly as
\begin{equation}
    \mathbb E_t
    \left[
        \min\left(
            \rho_t(\theta)\hat A_t,\,
            \operatorname{clip}(\rho_t(\theta),1-\epsilon,1+\epsilon)\hat A_t
        \right)
    \right].
\end{equation}
where $\rho_t(\theta)
    =
    \frac{
        p_\theta(\tilde a_t\mid s_t)
    }{
        p_{\theta_{\mathrm{old}}}(\tilde a_t\mid s_t)
    }$.
During training, this stochastic latent sampling is useful for exploration. At deployment, however, one often needs a deterministic controller. A choice is to set $z=0$, which is the
highest-density point of the Gaussian prior. However, this prior-centered inference rule is not always value-aware.

It is different from a standard Gaussian policy which the deterministic deployment action is usually the learned mean $\mu_\theta(s)$, and policy-gradient updates directly move this mean toward high-advantage actions. Therefore,
although the mean is not guaranteed to be globally optimal, it is explicitly optimized as the central
action of a unimodal policy. 
By contrast, in a flow-based policy, $z=0$ is only the highest-density point of the fixed latent prior. The actual action is obtained after a nonlinear transformation, this prior center is not guaranteed to correspond to the value-maximizing action in the executable action space.

This issue is especially important for GenPO because its augmented-action construction maps dummy actions to executable actions through
$a=\frac{x+y}{2}$. This projection is many-to-one: multiple dummy actions can induce the same executable action. Consequently, high likelihood in the augmented flow space does not uniquely determine which latent should be used for deterministic control.

\subsection{On-Policy Latent Steering Policy}
The above observation suggests a value-aware alternative to prior-centered inference.  Rather than using the fixed prior center $z=0$ as the deployment latent, we learn a state-conditioned Gaussian latent policy $\pi_\phi^Z(z\mid s)=\mathcal N(\mu_\phi(s),\Sigma)$, whose mean is optimized by reinforcement learning. At inference time, the learned mean $\mu_\phi(s)$ serves as the deterministic latent input for the trained GenPO policy. In this way, we transfer the useful deployment property of Gaussian policies to the latent space of a frozen flow-based policy: the deterministic inference latent is no longer the mode of a fixed prior, but the mean of a policy optimized for task return.

To this end, we freeze the trained GenPO policy $D_{\bar\theta}(s,z)=h(F_{\bar\theta}(s,z))
$, and reinterpret its latent input as the action of a new control problem. The latent policy selects
$z$, the frozen GenPO policy maps it to the executable action
$
    a=D_{\bar\theta}(s,z)
$, and the environment evolves under the original dynamics. This induces a latent-action MDP
$
    \mathcal M_{\bar\theta}^Z
    =
    (\mathcal S,\mathcal Z,P_{\bar\theta}^Z,r_{\bar\theta}^Z,\gamma)
$,
with
$
    P_{\bar\theta}^Z(s'\mid s,z)
    =
    P(s'\mid s,D_{\bar\theta}(s,z))
$,
and
$
    r_{\bar\theta}^Z(s,z)
    =
    r(s,D_{\bar\theta}(s,z))-\lambda_z\|z\|_2^2
$.
The latent regularization term keeps the learned latent policy near the reliable region of the GenPO prior and discourages querying out-of-prior latents.

We then optimize $\pi_\phi^Z$ with on-policy PPO in the induced latent-action MDP. Equivalently, the second-stage objective is
\begin{equation} 
 \mathbb E_t
    \left[
        \min\left(
            \rho_t^Z(\phi)\hat A_t^Z,\,
            \operatorname{clip}
            \left(
                \rho_t^Z(\phi),
                1-\epsilon,
                1+\epsilon
            \right)
            \hat A_t^Z
        \right)
    \right],
\end{equation}
where $ \rho_t^Z(\phi)=\frac{\pi_\phi^Z(z_t\mid s_t)}{\pi_{\phi_{\mathrm{old}}}^Z(z_t\mid s_t)}$.
Compared with the first-stage GenPO objective, the frozen flow policy is no longer updated. The optimization variable changes from the parameters of the generative policy to the parameters of the latent actor. Thus, the purpose of the second stage is not to reshape the entire action distribution,
but to learn a value-aware latent selection rule over the action manifold already learned by GenPO.

This latent-space formulation converts deterministic latent selection into an on-policy PPO problem. It preserves the training regime of GenPO and avoids introducing a replay buffer or an off-policy latent critic like DSRL. Instead of learning a latent action-value function $Q(s,z)$, we train a state-value function $V^Z(s)$ and compute advantages from on-policy rollouts in $\mathcal M_{\bar\theta}^Z$. Thus, policy gradients are taken only through the Gaussian latent actor, while the frozen GenPO policy is used solely for forward action generation.

\subsection{Practical Implementation}

We implement our method as a two-stage training pipeline. The first stage follows the practical GenPO implementation, and the second stage trains a lightweight latent actor on top of the frozen GenPO policy.
We train a GenPO policy in the original environment. Following GenPO, we include three practical components for stable on-policy training. First, we use the exact augmented-action likelihood to estimate the policy entropy and add an entropy regularization term, which encourages exploration. Second, we estimate the KL divergence between the current and old GenPO policies,
\begin{equation}
    \widehat D_{\mathrm{KL}}^{G}
    =
    \mathbb E_t
    \left[
        \log p_{\theta_{\mathrm{old}}}(\tilde a_t|s_t)
        -
        \log p_\theta(\tilde a_t|s_t)
    \right],
\end{equation}
and use it for KL-adaptive learning-rate control. Third, we use the dummy-action compression loss
\begin{equation}
    \mathcal L_{\mathrm{comp}}^{G}
    =
    \mathbb E_t\left[\|x_t-y_t\|_2^2\right],
\end{equation}
which reduces redundant exploration in the doubled action space and encourages the two dummy
components to represent consistent executable actions.



\begin{algorithm}[t]
    \caption{SteerGenPO}
    \label{alg:steergenpo}
    \begin{algorithmic}[1]
    \Statex \textbf{Stage I: Generative policy training}
    \State Initialize conditional flow $F_\theta$ and value network $V_\omega$.
    \For{each training iteration}
        \State Roll out with $\xi_t\sim p_0$, $\tilde a_t=(x_t,y_t)=F_\theta(s_t,\xi_t)$, and $a_t=h(\tilde a_t)=(x_t+y_t)/2$.
        \State Compute $\log p_\theta(\tilde a_t\mid s_t)$ by inverting $F_\theta$.
        \State Update $F_\theta,V_\omega$ using returns, likelihoods, entropy, and doubled-action consistency.
    \EndFor
    \State Freeze $F_\theta$ and define $D_\theta(s,z)=h(F_\theta(s,z))$.

    \Statex \textbf{Stage II: Latent steering policy training}
    \State Initialize $\pi^Z_\phi(z\mid s)=\mathcal{N}(\mu_\phi(s),\sigma_0^2 I)$ and $V^Z$; $\sigma_0$ sets the initial latent exploration scale.
    \For{each latent steering iteration}
        \State Sample $z_t\sim\pi^Z_\phi(\cdot\mid s_t)$ and execute $a_t=D_\theta(s_t,z_t)$ with $D_\theta$ fixed.
        \State Use the latent penalty to apply latent regularization: $\tilde r_t=r(s_t,a_t)-\lambda_z\|z_t\|_2^2$.
        \State Update only $\pi_\phi,V^Z$ using latent-action log probabilities and returns from $\tilde r_t$.
    \EndFor
    \Statex \textbf{Inference:} Use mean latent action $z_t=\mu_\phi(s_t)$ and execute $a_t=D_\theta(s_t,z_t)$ for deterministic steering.
    \end{algorithmic}
\end{algorithm}

\section{Experimental Results}
\label{sec:result}

\subsection{Experimental Setup}

We evaluate PPO, GenPO, and SteerGenPO in two settings. First, we use a six-task Isaac Lab~\cite{mittal2025isaac} benchmark covering Ant, Humanoid, Franka Arm, Anymal-D, Unitree Go2, and Unitree H1, spanning locomotion, legged control, and manipulation. Second, we evaluate Unitree G1 velocity tracking, transfer the learned policies to MuJoCo~\cite{todorov2012mujoco} for closed-loop command tracking, and deploy SteerGenPO on the real G1 robot. Implementation details and hyperparameters are provided in Appendix.
\begin{table}
    \centering
    \caption{Final return on Isaac Lab benchmark tasks. Mean $\pm$ standard deviation over 5 seeds.}
    \label{tab:final_return}
    \resizebox{\textwidth}{!}{
    \begin{tabular}{lcccccc}
        \toprule
        Method & Ant & Humanoid & Franka & Anymal-D & Unitree Go2 & Unitree H1 \\
        \midrule
        PPO & 145.79 $\pm$ 24.64 & 118.01 $\pm$ 41.08 & 114.61 $\pm$ 17.09 & 13.41 $\pm$ 7.09 & 18.66 $\pm$ 1.33 & 26.31 $\pm$ 0.56 \\
        GenPO & 171.76 $\pm$ 14.67 & 196.64 $\pm$ 15.12 & 147.28 $\pm$ 1.06 & 20.63 $\pm$ 0.14 & 26.69 $\pm$ 0.45 & 24.96 $\pm$ 0.25 \\
        SteerGenPO & \textbf{195.94 $\pm$ 14.17} & \textbf{222.53 $\pm$ 23.24} & \textbf{150.44 $\pm$ 2.35} & \textbf{24.12 $\pm$ 0.16} & \textbf{28.65 $\pm$ 0.54} & \textbf{27.65 $\pm$ 0.63} \\
        \bottomrule
    \end{tabular}}
\end{table}

\subsection{Benchmark evaluation}

\begin{figure}                                                             
    \centering
    \includegraphics[width=\linewidth]{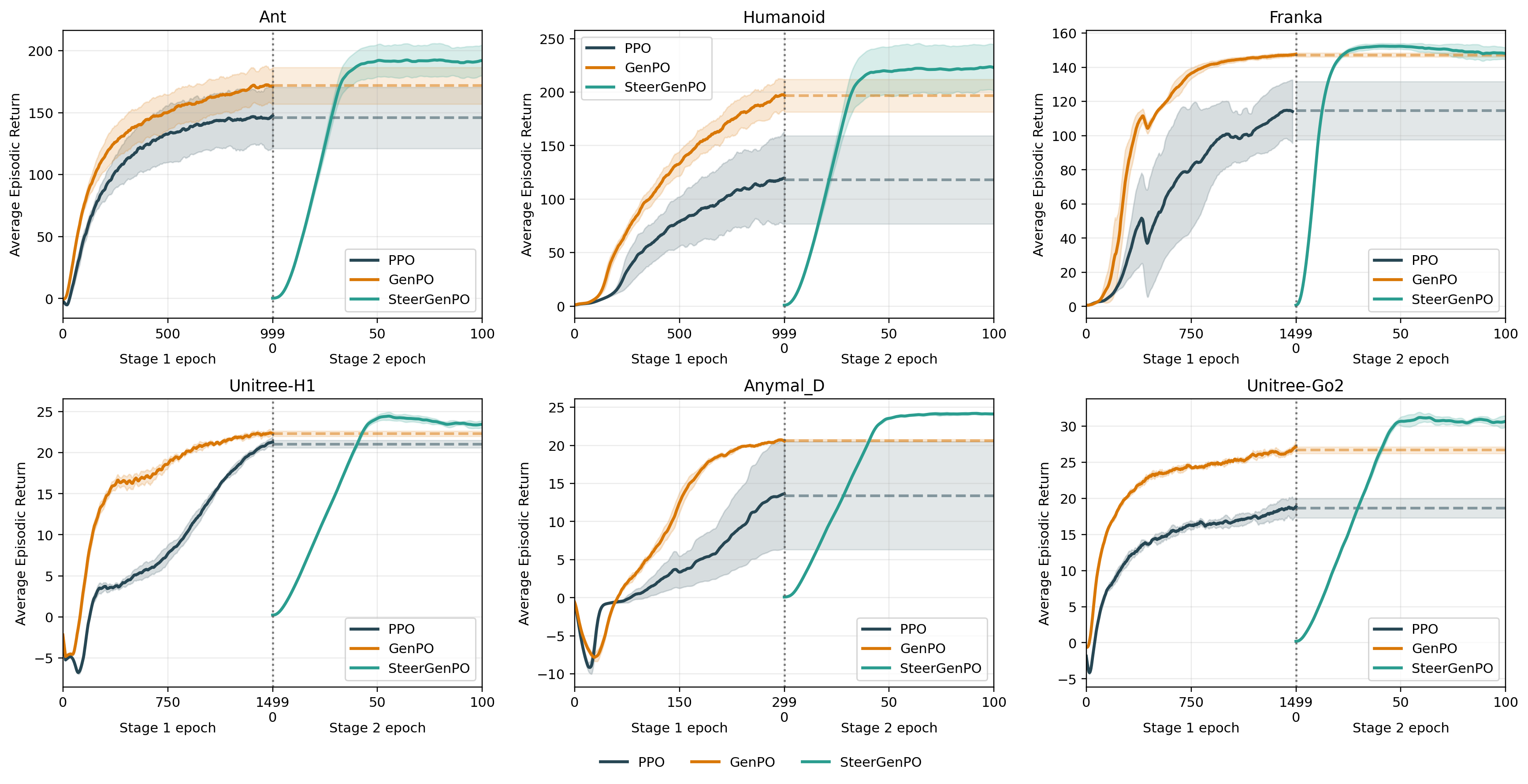}          
    \caption{
  Two-stage training curves on six Isaac Lab tasks. Stage I trains PPO and GenPO from scratch; Stage II freezes the converged GenPO generator and trains only the SteerGenPO latent actor. Dashed lines indicate final Stage-I returns. Curves show mean $\pm$ standard deviation over seeds.
    }
    \label{fig:benchmark_latent_curves}                                       
\end{figure}

Figure~\ref{fig:benchmark_latent_curves} summarizes the training behavior of PPO, GenPO, and SteerGenPO on the six Isaac Lab benchmark tasks. SteerGenPO reaches higher final returns than both baselines across tasks. GenPO provides a strong generative action prior in the first stage, while the second-stage latent actor further improves the frozen generator with relatively few updates. This shows that the learned generator induces a useful latent action space where state-conditioned latent selection can recover stronger behaviors.

Table~\ref{tab:final_return} reports final returns. SteerGenPO achieves the best performance on all six tasks, with the largest gains on Humanoid, Anymal-D, and Unitree , indicating that high-dimensional control benefits most from combining generative action priors with latent steering. The lower standard deviations on most tasks also suggest improved reproducibility.


Additional ablations are provided in Appendix. They show that latent selection is critical even when the GenPO generator is fixed: random latents are less stable, the zero latent is a strong deterministic baseline, and the learned state-conditioned latent actor performs best. The ablations also show that the initial latent exploration scale must be chosen carefully to balance adaptation and staying within the reliable operating region of the frozen generator.

Overall, SteerGenPO improves final return over PPO by up to $88.6\%$ and over GenPO by up to $16.9\%$, demonstrating that latent steering turns a stochastic generative policy into a stronger deterministic controller.

\subsection{Humanoid Locomotion Evaluation}

We further evaluate SteerGenPO on the Unitree G1 with the velocity-tracking task. SteerGenPO achieves higher training returns than PPO and GenPO, showing that latent steering remains effective on higher-dimensional humanoid control. Training curves are provided in Appendix. We then evaluate whether this improvement translates to closed-loop path following.

\paragraph{Closed-loop Path Following in MuJoCo}


In trajectory tracking, a shared high-level planner converts root-pose error into body-frame velocity commands $(v_x, v_y, \omega_z)$, which are executed by each learned policy. We test triangle, figure-eight, square, S-curve, and zigzag paths, covering sharp turns, repeated reorientation, lateral correction, and yaw control.

\begin{figure}[t]                                         
      \centering                                            
      \includegraphics[width=0.9\linewidth]{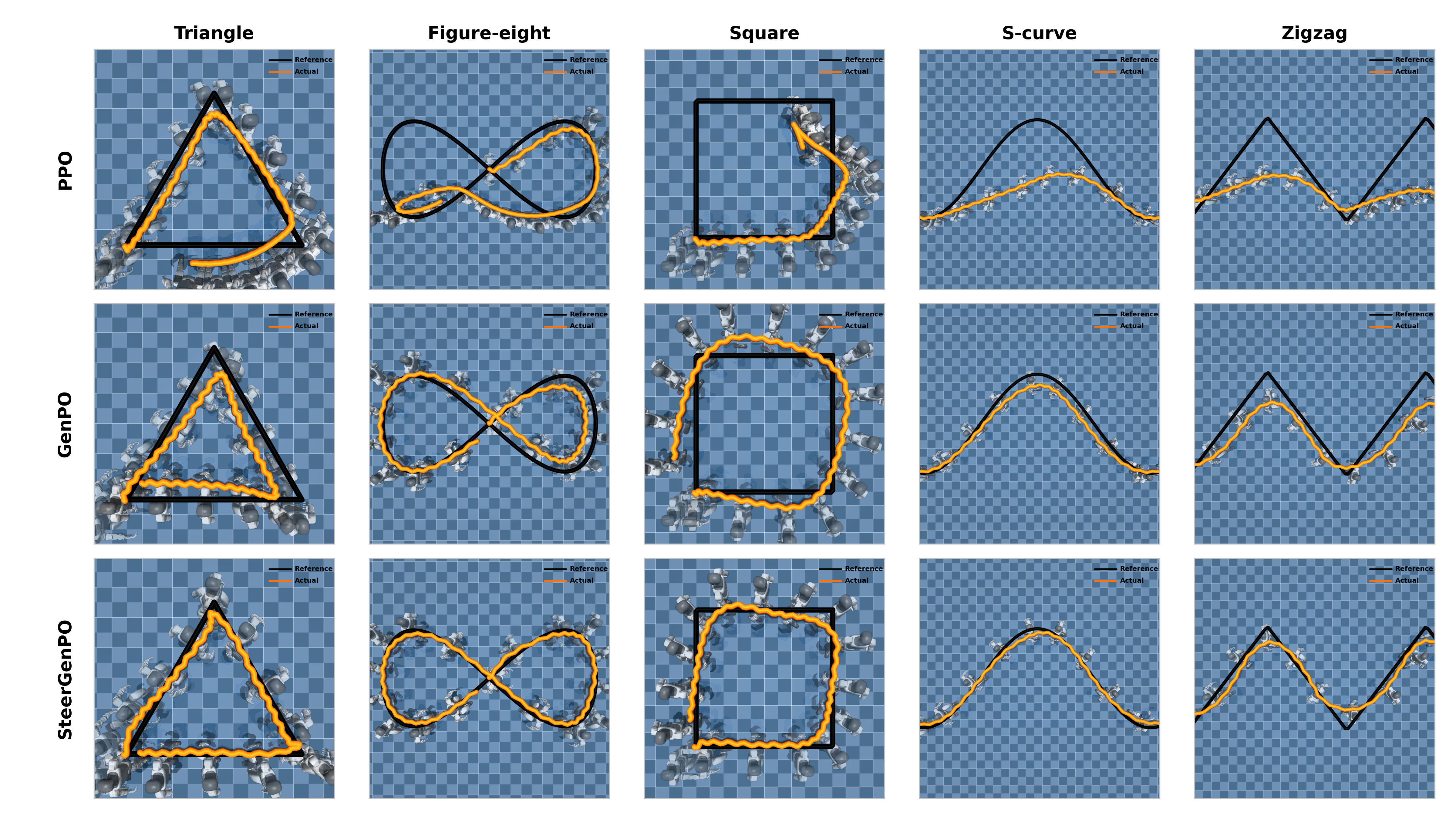}             
      \caption{                                    
        Closed-loop command tracking on Unitree G1 in MuJoCo.     
      Rows show policies and columns show reference trajectories.          
      Black curves are predefined reference paths, colored curves are realized root trajectories, and robot snapshots are overlaid every 2 seconds.               
      }                                             
      \label{fig:tracking}
  \end{figure}

Figure~\ref{fig:tracking} shows that, under the same planner, SteerGenPO tracks the reference paths more reliably than other methods. On the figure-eight path, GenPO and SteerGenPO progress more uniformly, whereas PPO covers a smaller portion after repeated heading corrections. At sharp turns, SteerGenPO more consistently reorients before advancing, indicating more accurate yaw-command execution.

These results show that latent steering improves the stability and precision of generative locomotion policies under structured, time-varying commands.

\begin{figure}
    \centering
    \includegraphics[width=0.8\linewidth]{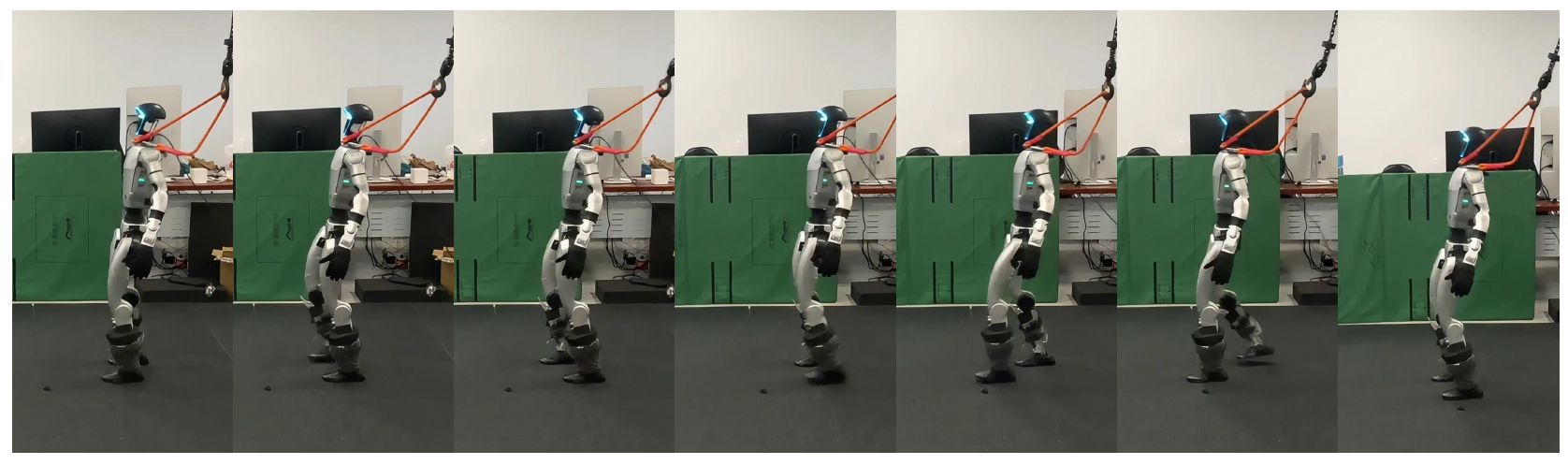}
    \caption{Real-world Unitree G1 walking with the learned SteerGenPO controller.}
    \label{fig:walking}
\end{figure}

\paragraph{Real-world Deployment}                       
  We further deploy the SteerGenPO controller on the real Unitree G1 robot to verify its sim-to-real feasibility. As shown in Figure~\ref{fig:walking}, the learned controller produces stable forward walking on hardware, with consistent whole-body coordination and no additional real-world fine-tuning. This qualitative deployment result suggests that latent steering preserves the deployable motion prior learned in simulation while improving the reliability of the final controller. 

  

\section{Limitation}
Despite its empirical effectiveness, SteerGenPO has several limitations. It relies on the quality and coverage of the first-stage generative policy, since latent steering cannot recover behaviors outside the learned action manifold. Although the second-stage latent training is substantially cheaper than fine-tuning the full generative policy, it still introduces an additional optimization step before deployment. Future work will study more sample-efficient latent policy learning and extend latent steering to broader generative-policy settings and robotic tasks.

\section{Conclusion}
\label{clusion}
We presented SteerGenPO, a latent-space reinforcement learning framework for steering generative policies into stable controllers. The central idea is to convert stochastic latent sampling into deterministic state-conditioned latent steering. During training, stochastic generative sampling provides diverse action proposals and helps discover high-return behaviors. At test time, random latent sampling is replaced by a learned latent actor, yielding a deterministic controller that preserves the expressive action manifold learned by the generative policy while producing more reproducible actions. SteerGenPO improves over both PPO and GenPO policies. The resulting controller achieves higher returns and efficient second-stage adaptation. In closed-loop path-following tasks, SteerGenPO also produces more accurate command responses and more stable trajectory tracking performance. These results suggest that latent steering is an effective way to convert stochastic generative policies into deterministic controllers.


\section*{Acknowledgment}
This work was supported by the National Natural Science Foundation of China (62303319, 62406195), HPC Platform of ShanghaiTech University, and MoE Key Laboratory of Intelligent Perception and Human-Machine Collaboration (ShanghaiTech University), Shanghai Engineering Research Center of Intelligent Vision and Imaging. This work was also supported in part by computational resources provided by Fcloud CO., LTD.



\bibliography{reference}  

\clearpage
\appendix
\section{Isaaclab Benchmark Experiments}
We provide training hyperparameters, network architectures, and task-specific settings used in the experiments.

\subsection{Environmental Details}

We evaluate all methods on six continuous-control tasks in IsaacLab: Isaac-Ant-v0, Isaac-Humanoid-v0, Isaac-Lift-Cube-Franka-v0, Isaac-Velocity-Flat-Anymal-D-v0, Isaac-Velocity-Rough-Unitree-Go2-v0, and Isaac-Velocity-Rough-H1-v0. These tasks cover classic locomotion, robotic manipulation, and legged locomotion on both flat and rough terrain. The corresponding observation and action dimensions are $(60,8)$, $(87,21)$, $(36,8)$, $(48,12)$, $(235,12)$, and $(256,19)$, respectively. All training runs use 2048 parallel environments.

\subsection{Hardware Configurations}

All experiments were carried out on a server equipped with two Intel Xeon Gold 5218R CPUs and 8 NVIDIA GeForce RTX 4090 GPUs. Each CPU contains 20 physical cores and supports 2 threads per core, yielding 40 physical cores and 80 logical CPUs in total. The base frequency of the CPUs is 2.10 GHz, with a maximum frequency of 4.00 GHz. The system supports 46-bit physical and 48-bit virtual addressing and is configured with a two-node NUMA topology. The server has 503 GiB of system memory and 2 GiB of swap space.

For GPU acceleration, we used 8 NVIDIA GeForce RTX 4090 GPUs, each equipped with 24 GB of memory and connected via PCIe. The GPUs were operated with NVIDIA driver version 575.64.03 and CUDA 12.9. GPU acceleration was used for IsaacLab simulation and neural network optimization. No ECC was enabled during the experiments, and MIG was not available on the GPUs. All methods were evaluated on the same hardware platform.

\subsection{Reinforcement Learning Framework in IsaacLab}

IsaacLab supports multiple reinforcement learning backends, including RSL-RL, RL-Games, SKRL, and Stable-Baselines3. In our experiments, all algorithms are implemented and trained using RSL-RL version 3.1.2. We use the same IsaacLab task interface and the same RSL-RL training pipeline for PPO, GenPO, and SteerGenPO, which avoids differences caused by using different reinforcement learning libraries.

PPO is used as the standard on-policy actor-critic baseline. GenPO follows the same on-policy training protocol while replacing the policy representation with a generative policy model. SteerGenPO builds on a pretrained GenPO policy. During SteerGenPO training, the GenPO policy is frozen, and a latent PPO policy is trained to produce latent actions that are mapped to environment actions through the frozen GenPO policy. This setup keeps the environment interface and training pipeline consistent across all methods.

\subsection{Hyperparameters}

The hyperparameter configurations are reported in the appendix tables. For each task, we provide one table comparing PPO, GenPO, and SteerGenPO. The tables include the network architecture, activation function, number of parallel environments, rollout length, GAE parameters, discount factor, learning rates, gradient clipping threshold, entropy coefficient, value loss coefficient, desired KL, and GenPO-specific generative policy parameters. Parameters that are not applicable to PPO are denoted by ``/''.

For SteerGenPO, the main task-level table reports the configuration same as GenPO. The additional SteerGenPO table reports the additional hyperparameters, including latent actor and critic architectures, activation function, initial action standard deviation, learning rate, number of learning epochs, number of mini-batches, entropy coefficient, value loss coefficient, desired KL, and latent regularization strength.

\begin{figure}[ht]
    \centering
    \includegraphics[width=\linewidth]{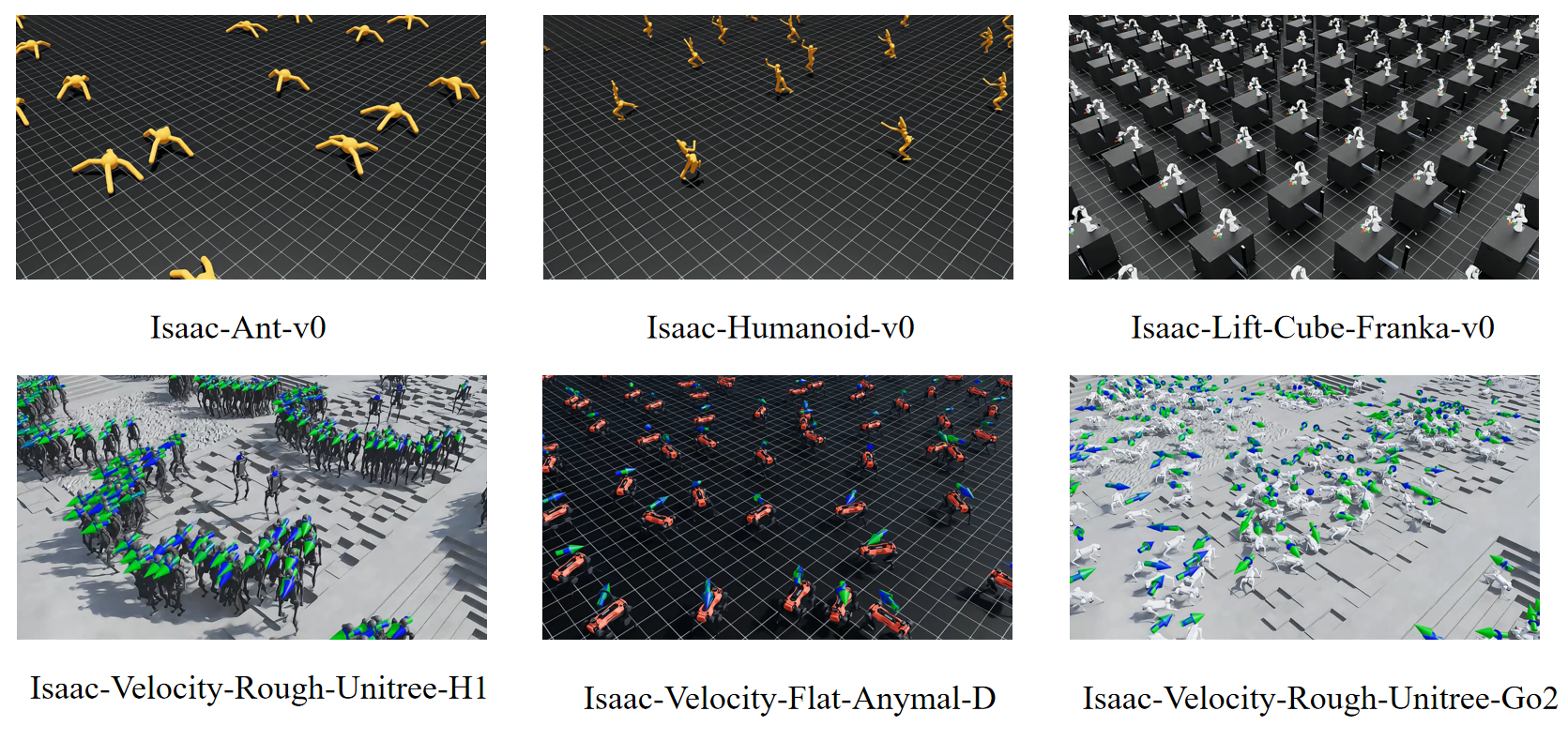}
    \caption{
    Representative rollout snapshots of the trained SteerGenPO policies on the six Isaac Lab benchmark tasks.
    The benchmarks cover diverse control settings, including locomotion, humanoid control, legged robot control,
    and robotic manipulation. These qualitative results complement the quantitative benchmark results reported in
    the main paper.
    }
    \label{fig:benchmark_play}
\end{figure}

\begin{table}[ht]
  \centering
  \caption{Hyper-parameters used in Isaaclab-Ant-v0.}
  \label{tab:hyperparams_ant}
  \resizebox{\textwidth}{!}{
  \begin{tabular}{lccc}
  \toprule
  Hyperparameter & PPO & GenPO & SteerGenPO \\
  \midrule
  Hidden layers in actor network & {[400,200,100]} & {[400,200,100]} & {[400,200,100]} \\
  Hidden layers in critic network & {[400,200,100]} & {[400,200,100]} & {[400,200,100]} \\
  Activation & elu & mish & mish \\
  Number of environments & 2048 & 2048 & 2048 \\
  Rollout length & 32 & 32 & 32 \\
  Use GAE & True & True & True \\
  Discount for reward $\gamma$ & 0.99 & 0.99 & 0.99 \\
  GAE smoothing parameter $\lambda$ & 0.95 & 0.95 & 0.95 \\
  Learning rate for actor & $1 \times 10^{-3}$ & $1 \times 10^{-3}$ & $1 \times 10^{-3}$ \\
  Learning rate for critic & $1 \times 10^{-3}$ & $1 \times 10^{-3}$ & $1 \times 10^{-3}$ \\
  Actor critic grad norm & 1.0 & 1.0 & 1.0 \\
  Entropy coefficient & 0.01 & 0.01 & 0.01 \\
  Value loss coefficient & 1.0 & 1.0 & 1.0 \\
  Desired KL & 0.01 & 0.01 & 0.01 \\
  Diffusion steps & {--} & 5 & 5 \\
  Compress coefficient $\lambda_c$ & {--} & 0.01 & 0.01 \\
  Time embedding dimension & {--} & 32 & 32 \\
  Hidden layers in time embedding & {--} & {[256,256]} & {[256,256]} \\
  Mixing coefficient $p$ & {--} & 0.95 & 0.95 \\
  Diffusion std & {--} & 3.0 & 3.0 \\
  \bottomrule
  \end{tabular}
  }
\end{table}

\begin{table}[ht]
  \centering
  \caption{Hyper-parameters used in Isaaclab-Humanoid-v0.}
  \label{tab:hyperparams_humanoid}
  \resizebox{\textwidth}{!}{
  \begin{tabular}{lccc}
  \toprule
  Hyperparameter & PPO & GenPO & SteerGenPO \\
  \midrule
  Hidden layers in actor network & {[400,200,100]} & {[400,200,100]} & {[400,200,100]} \\
  Hidden layers in critic network & {[400,200,100]} & {[400,200,100]} & {[400,200,100]} \\
  Activation & elu & mish & mish \\
  Number of environments & 2048 & 2048 & 2048 \\
  Rollout length & 32 & 32 & 32 \\
  Use GAE & True & True & True \\
  Discount for reward $\gamma$ & 0.99 & 0.99 & 0.99 \\
  GAE smoothing parameter $\lambda$ & 0.95 & 0.95 & 0.95 \\
  Learning rate for actor & $5 \times 10^{-4}$ & $5 \times 10^{-4}$ & $5 \times 10^{-4}$ \\
  Learning rate for critic & $5 \times 10^{-4}$ & $5 \times 10^{-4}$ & $5 \times 10^{-4}$ \\
  Actor critic grad norm & 1.0 & 1.0 & 1.0 \\
  Entropy coefficient & 0.0 & 0.0 & 0.0 \\
  Value loss coefficient & 2.0 & 2.0 & 2.0 \\
  Desired KL & 0.01 & 0.01 & 0.01 \\
  Diffusion steps & {--} & 5 & 5 \\
  Compress coefficient $\lambda_c$ & {--} & 0.01 & 0.01 \\
  Time embedding dimension & {--} & 32 & 32 \\
  Hidden layers in time embedding & {--} & {[256,256]} & {[256,256]} \\
  Mixing coefficient $p$ & {--} & 0.9 & 0.9 \\
  Diffusion std & {--} & 3.0 & 3.0 \\
  \bottomrule
  \end{tabular}
  }
\end{table}

\begin{table}[ht]
  \centering
  \caption{Hyper-parameters used in Isaaclab-Franka-Arm-v0.}
  \label{tab:hyperparams_franka}
  \resizebox{\textwidth}{!}{
  \begin{tabular}{lccc}
  \toprule
  Hyperparameter & PPO & GenPO & SteerGenPO \\
  \midrule
  Hidden layers in actor network & {[256,128,64]} & {[256,128,64]} & {[256,128,64]} \\
  Hidden layers in critic network & {[256,128,64]} & {[256,128,64]} & {[256,128,64]} \\
  Activation & elu & mish & mish \\
  Number of environments & 2048 & 2048 & 2048 \\
  Rollout length & 24 & 24 & 24 \\
  Use GAE & True & True & True \\
  Discount for reward $\gamma$ & 0.98 & 0.98 & 0.98 \\
  GAE smoothing parameter $\lambda$ & 0.95 & 0.95 & 0.95 \\
  Learning rate for actor & $1 \times 10^{-4}$ & $1 \times 10^{-4}$ & $1 \times 10^{-4}$ \\
  Learning rate for critic & $1 \times 10^{-4}$ & $1 \times 10^{-4}$ & $1 \times 10^{-4}$ \\
  Actor critic grad norm & 1.0 & 1.0 & 1.0 \\
  Entropy coefficient & 0.006 & 0.006 & 0.006 \\
  Value loss coefficient & 1.0 & 1.0 & 1.0 \\
  Desired KL & 0.01 & 0.01 & 0.01 \\
  Diffusion steps & {--} & 5 & 5 \\
  Compress coefficient $\lambda_c$ & {--} & 0.01 & 0.01 \\
  Time embedding dimension & {--} & 32 & 32 \\
  Hidden layers in time embedding & {--} & {[256,256]} & {[256,256]} \\
  Mixing coefficient $p$ & {--} & 0.9 & 0.9 \\
  Diffusion std & {--} & 3.0 & 3.0 \\
  \bottomrule
  \end{tabular}
  }
\end{table}

\begin{table}[ht]
  \centering
  \caption{Hyper-parameters used in Isaaclab-Anymal-D-v0.}
  \label{tab:hyperparams_anymal_d}
  \resizebox{\textwidth}{!}{
  \begin{tabular}{lccc}
  \toprule
  Hyperparameter & PPO & GenPO & SteerGenPO \\
  \midrule
  Hidden layers in actor network & {[128,128,128]} & {[128,128,128]} & {[128,128,128]} \\
  Hidden layers in critic network & {[128,128,128]} & {[128,128,128]} & {[128,128,128]} \\
  Activation & elu & mish & mish \\
  Number of environments & 2048 & 2048 & 2048 \\
  Rollout length & 24 & 24 & 24 \\
  Use GAE & True & True & True \\
  Discount for reward $\gamma$ & 0.99 & 0.99 & 0.99 \\
  GAE smoothing parameter $\lambda$ & 0.95 & 0.95 & 0.95 \\
  Learning rate for actor & $1 \times 10^{-3}$ & $1 \times 10^{-3}$ & $1 \times 10^{-3}$ \\
  Learning rate for critic & $1 \times 10^{-3}$ & $1 \times 10^{-3}$ & $1 \times 10^{-3}$ \\
  Actor critic grad norm & 1.0 & 1.0 & 1.0 \\
  Entropy coefficient & 0.005 & 0.005 & 0.005 \\
  Value loss coefficient & 1.0 & 1.0 & 1.0 \\
  Desired KL & 0.01 & 0.01 & 0.01 \\
  Diffusion steps & {--} & 5 & 5 \\
  Compress coefficient $\lambda_c$ & {--} & 0.01 & 0.01 \\
  Time embedding dimension & {--} & 32 & 32 \\
  Hidden layers in time embedding & {--} & {[256,256]} & {[256,256]} \\
  Mixing coefficient $p$ & {--} & 0.95 & 0.95 \\
  Diffusion std & {--} & 3.0 & 3.0 \\
  \bottomrule
  \end{tabular}
  }
\end{table}

\begin{table}[ht]
  \centering
  \caption{Hyper-parameters used in Isaaclab-Unitree-Go2-v0.}
  \label{tab:hyperparams_unitree_go2}
  \resizebox{\textwidth}{!}{
  \begin{tabular}{lccc}
  \toprule
  Hyperparameter & PPO & GenPO & SteerGenPO \\
  \midrule
  Hidden layers in actor network & {[512,256,128]} & {[512,256,128]} & {[512,256,128]} \\
  Hidden layers in critic network & {[512,256,128]} & {[512,256,128]} & {[512,256,128]} \\
  Activation & elu & mish & mish \\
  Number of environments & 2048 & 2048 & 2048 \\
  Rollout length & 24 & 24 & 24 \\
  Use GAE & True & True & True \\
  Discount for reward $\gamma$ & 0.99 & 0.99 & 0.99 \\
  GAE smoothing parameter $\lambda$ & 0.95 & 0.95 & 0.95 \\
  Learning rate for actor & $1 \times 10^{-3}$ & $1 \times 10^{-3}$ & $1 \times 10^{-3}$ \\
  Learning rate for critic & $1 \times 10^{-3}$ & $1 \times 10^{-3}$ & $1 \times 10^{-3}$ \\
  Actor critic grad norm & 1.0 & 1.0 & 1.0 \\
  Entropy coefficient & 0.01 & 0.01 & 0.01 \\
  Value loss coefficient & 1.0 & 1.0 & 1.0 \\
  Desired KL & 0.01 & 0.01 & 0.01 \\
  Diffusion steps & {--} & 5 & 5 \\
  Compress coefficient $\lambda_c$ & {--} & 0.01 & 0.01 \\
  Time embedding dimension & {--} & 32 & 32 \\
  Hidden layers in time embedding & {--} & {[256,256]} & {[256,256]} \\
  Mixing coefficient $p$ & {--} & 0.9 & 0.9 \\
  Diffusion std & {--} & 3.0 & 3.0 \\
  \bottomrule
  \end{tabular}
  }
\end{table}

\begin{table}[ht]
  \centering
  \caption{Hyper-parameters used in Isaaclab-Unitree-H1-v0.}
  \label{tab:hyperparams_unitree_h1}
  \resizebox{\textwidth}{!}{
  \begin{tabular}{lccc}
  \toprule
  Hyperparameter & PPO & GenPO & SteerGenPO \\
  \midrule
  Hidden layers in actor network & {[512,256,128]} & {[512,256,128]} & {[512,256,128]} \\
  Hidden layers in critic network & {[512,256,128]} & {[512,256,128]} & {[512,256,128]} \\
  Activation & elu & mish & mish \\
  Number of environments & 2048 & 2048 & 2048 \\
  Rollout length & 24 & 24 & 24 \\
  Use GAE & True & True & True \\
  Discount for reward $\gamma$ & 0.99 & 0.99 & 0.99 \\
  GAE smoothing parameter $\lambda$ & 0.95 & 0.95 & 0.95 \\
  Learning rate for actor & $1 \times 10^{-3}$ & $1 \times 10^{-3}$ & $1 \times 10^{-3}$ \\
  Learning rate for critic & $1 \times 10^{-3}$ & $1 \times 10^{-3}$ & $1 \times 10^{-3}$ \\
  Actor critic grad norm & 1.0 & 1.0 & 1.0 \\
  Entropy coefficient & 0.01 & 0.01 & 0.01 \\
  Value loss coefficient & 1.0 & 1.0 & 1.0 \\
  Desired KL & 0.01 & 0.01 & 0.01 \\
  Diffusion steps & {--} & 5 & 5 \\
  Compress coefficient $\lambda_c$ & {--} & 0.01 & 0.01 \\
  Time embedding dimension & {--} & 32 & 32 \\
  Hidden layers in time embedding & {--} & {[256,256]} & {[256,256]} \\
  Mixing coefficient $p$ & {--} & 0.9 & 0.9 \\
  Diffusion std & {--} & 3.0 & 3.0 \\
  \bottomrule
  \end{tabular}
  }
\end{table}

\begin{table}[ht]
  \centering
  \caption{Additional hyperparameters used in SteerGenPO.}
  \label{tab:hyperparams_steergenpo_extra}
  \resizebox{\textwidth}{!}{
  \begin{tabular}{lcccccc}
  \toprule
  Hyperparameter & Ant & Humanoid & Franka Arm & Anymal-D & Unitree-Go2 & Unitree-H1 \\
  \midrule
  Latent actor hidden layers & {[400,200,100]} & {[400,200,100]} & {[512,256,128]} & {[128,128,128]} & {[512,256,128]} & {[1024,512,256]} \\
  Latent critic hidden layers & {[400,200,100]} & {[400,200,100]} & {[512,256,128]} & {[128,128,128]} & {[512,256,128]} & {[1024,512,256]} \\
  Latent activation & elu & elu & elu & elu & elu & elu \\
  Latent initial action std & 0.01 & 0.01 & 0.01 & 0.01 & 0.001 & 0.001 \\
  Latent learning epochs & 3 & 3 & 3 & 3 & 3 & 3 \\
  Latent mini-batches & 4 & 4 & 4 & 4 & 4 & 4 \\
  Latent learning rate & $2 \times 10^{-4}$ & $5 \times 10^{-4}$ & $3 \times 10^{-4}$ & $3 \times 10^{-4}$ & $2 \times 10^{-4}$ & $2 \times 10^{-4}$ \\
  Latent entropy coefficient & 0.003 & 0.0 & 0.008 & 0.005 & 0.008 & 0.008 \\
  Latent value loss coefficient & 1.0 & 1.0 & 1.0 & 1.0 & 1.0 & 1.0 \\
  Latent desired KL & 0.01 & 0.005 & 0.005 & 0.005 & 0.005 & 0.005 \\
  Latent penalty strength & 0.001 & 0.001 & 0.001& 0.001 & 0.001 &0.001\\
  \bottomrule
  \end{tabular}
  }
\end{table}

\section{Ablation Studies}
\label{app:ablations}

We conduct two ablations to study how deployment-time latent inputs affect a fixed, frozen GenPO generator. 

\subsection{Latent Input Ablation}
First, for each task, we fix a representative GenPO base policy and compare three ways of choosing the latent input: a randomly sampled latent, a fixed zero latent, and our learned state-conditioned latent actor. All three variants use the same underlying generator $g_\theta$,  so the differences in Table~\ref{tab:performance} reflect the effect of latent-space guidance rather than changes in the base policy. 

This controlled comparison shows that the deployment-time latent choice has a large effect even when the generator is fixed.  Random latent sampling gives lower returns and less consistent behavior, indicating that the stochasticity useful for exploration during training is not directly suitable for deployment. Zero-latent deployment is also a strong deterministic baseline, since the converged GenPO generator often maps the center of the latent prior to stable high-return actions.

\begin{table}[ht]
\centering
\caption{Latent-selection ablation on frozen GenPO policies. Mean return $\pm$ standard deviation over five evaluation seeds.}
\label{tab:performance}
\resizebox{\textwidth}{!}{
\begin{tabular}{lcccccc}
\toprule
Method & Ant & Anymal\_D & Franka & Humanoid & Unitree-Go2 & Unitree-H1 \\
\midrule
GenPO-random & $160.93 \pm 2.94$ & $19.37 \pm 0.53$ & $154.22 \pm 0.62$ & $119.87 \pm 9.54$ & $24.55 \pm 0.57$ & $26.13 \pm 0.42$ \\
GenPO-zero & $171.92 \pm 3.27$ & $23.20 \pm 0.40$ & $155.61 \pm 0.12$ & $192.76 \pm 5.14$ & $27.09 \pm 0.25$ & $28.69 \pm 0.67$ \\
SteerGenPO & $\mathbf{179.34 \pm 2.81}$ & $\mathbf{23.56 \pm 0.35}$ & $\mathbf{157.53 \pm 0.37}$ & $\mathbf{206.53 \pm 1.66}$ & $\mathbf{27.23 \pm 0.47}$ & $\mathbf{29.57 \pm 0.41}$ \\
\bottomrule
\end{tabular}
}
\end{table}

\subsection{Latent Initial Standard Deviation Ablation}
\label{app:initstd_ablation}
\begin{figure}[ht]
    \centering
    \includegraphics[width=0.5\linewidth]{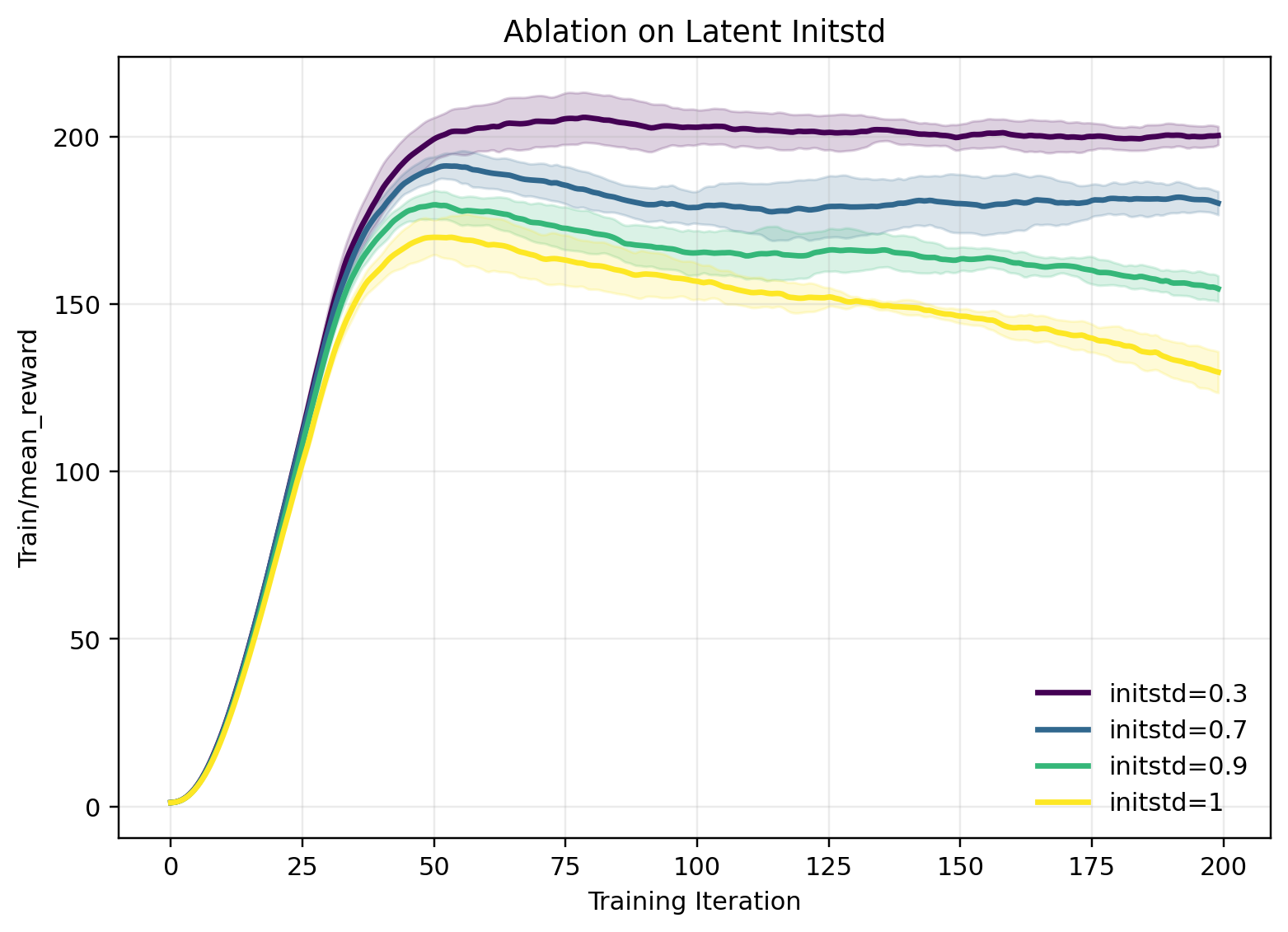}
    \caption{Ablation study on the standard deviation (std) of latent initialization.}
    \label{fig:std}
\end{figure}

Ablation results in Figure~\ref{fig:std} on the initial standard deviation of the policy exploration noise confirms that too little noise leads to insufficient exploration, while excessive noise destabilizes training; the best performance is achieved at a moderate value $0.3$ that balances these two factors.

\section{Unitree G1 Walking Experiment}

\subsection{Training Framework}

We implement the Unitree G1 walking task based on the Unitree RL Lab framework. Unitree RL Lab provides a modular reinforcement learning pipeline for Unitree legged robots, including robot configuration, actuator modeling, observation construction, reward and termination managers, command sampling, curriculum learning, and domain randomization. The framework is built around massively parallel GPU simulation and allows us to train thousands of environments simultaneously for locomotion policy learning.

For policy optimization, we use RSL-RL 3.1.2 as the reinforcement learning backend. PPO is trained with the standard on-policy actor-critic runner, while GenPO extends the same training pipeline with a flow-based policy parameterization. SteerGenPO is trained in two stages: the first stage follows the GenPO configuration, and the second stage performs latent-space fine-tuning using a PPO-style runner with additional latent-action control parameters. This implementation choice keeps the environment, observation space, reward definition, and rollout interface consistent across all methods, so that the comparison mainly reflects differences in policy parameterization and optimization strategy.

\subsection{Hardware Setup}

All training experiments are conducted on a workstation equipped with eight NVIDIA GeForce RTX 4090 GPUs, each with 48 GB of memory. The machine uses an Intel Xeon Platinum 8368Q CPU at 2.60 GHz with 152 logical CPU cores and 503 GB of system memory. The NVIDIA driver version is 570.86.10 and the CUDA version is 12.8.

Each training run uses two RTX 4090 GPUs. We launch 4096 parallel simulated environments on each GPU, resulting in 8192 environments in total for one training run. This high degree of parallelism is used to accelerate on-policy data collection and stabilize policy optimization for the Unitree G1 walking task.

\subsection{Reward Function Design}

We employ a largely shared reward design across PPO, GenPO, and SteerGenPO. All three methods optimize the same set of locomotion-oriented reward terms, including velocity tracking rewards, survival reward, body stabilization penalties, joint regularization penalties, gait-shaping rewards, and contact-related penalties. This design keeps the task objective consistent across algorithms and isolates the effect of the policy parameterization and optimization strategy.

More specifically, the reward consists of 19 terms: horizontal linear velocity tracking, yaw-rate tracking, alive bonus, vertical base velocity penalty, roll/pitch angular velocity penalty, joint velocity penalty, joint acceleration penalty, action-rate penalty, joint-limit penalty, energy penalty, arm-joint deviation penalty, waist-joint deviation penalty, leg-joint deviation penalty, base orientation penalty, base-height penalty, gait reward, foot-sliding penalty, foot-clearance reward, and undesired-contact penalty. The total reward is computed as a weighted sum of these terms:
\begin{equation}
    r_t = \sum_i w_i \mathcal{R}_i,
\end{equation}
where $\mathcal{R}_i$ denotes the $i$-th reward term and $w_i$ denotes its coefficient.

Table~\ref{tab:reward_terms} reports the reward formulae and coefficients used for SteerGenPO training. PPO and GenPO use the same reward terms. The coefficients are shared across algorithms. 

\begin{table*}[ht]
\centering
\caption{Reward terms, formulae, and coefficients used for SteerGenPO training. PPO and GenPO use the same reward terms and coefficient values.}
\label{tab:reward_terms}
\resizebox{\textwidth}{!}{
\begin{tabular}{lp{8.2cm}c}
\toprule
Reward term & Formula & Coef. \\
\midrule

$\mathcal{R}_{\mathrm{track\_lin\_vel\_xy}}$
& $\exp\!\left(-\frac{\|\mathbf{v}^{\mathrm{cmd}}_{xy}-\mathbf{v}_{xy}\|_2^2}{0.5^2}\right)$
& 1.0 \\

$\mathcal{R}_{\mathrm{track\_ang\_vel\_z}}$
& $\exp\!\left(-\frac{(\omega^{\mathrm{cmd}}_z-\omega_z)^2}{0.5^2}\right)$
& 0.5 \\

$\mathcal{R}_{\mathrm{alive}}$
& $1$
& 0.15 \\

$\mathcal{R}_{\mathrm{base\_linear\_velocity}}$
& $v_z^2$
& -2.0 \\

$\mathcal{R}_{\mathrm{base\_angular\_velocity}}$
& $\|\boldsymbol{\omega}_{xy}\|_2^2$
& -0.05 \\

$\mathcal{R}_{\mathrm{joint\_vel}}$
& $\|\dot{\mathbf{q}}\|_2^2$
& -0.001 \\

$\mathcal{R}_{\mathrm{joint\_acc}}$
& $\|\ddot{\mathbf{q}}\|_2^2$
& $-2.5\times 10^{-7}$ \\

$\mathcal{R}_{\mathrm{action\_rate}}$
& $\|\mathbf{a}_t-\mathbf{a}_{t-1}\|_2^2$
& -0.01 \\

$\mathcal{R}_{\mathrm{dof\_pos\_limits}}$
& $\sum_i \left[\max(q_i-q_i^{\max},0)+\max(q_i^{\min}-q_i,0)\right]$
& -5.0 \\

$\mathcal{R}_{\mathrm{energy}}$
& $\sum_i |\tau_i|\,|\dot{q}_i|$
& $-2.0\times 10^{-5}$ \\

$\mathcal{R}_{\mathrm{joint\_deviation\_arms}}$
& $\|\mathbf{q}_{\mathrm{arms}}-\mathbf{q}^{\mathrm{def}}_{\mathrm{arms}}\|_1$
& -0.1 \\

$\mathcal{R}_{\mathrm{joint\_deviation\_waists}}$
& $\|\mathbf{q}_{\mathrm{waist}}-\mathbf{q}^{\mathrm{def}}_{\mathrm{waist}}\|_1$
& -1.0 \\

$\mathcal{R}_{\mathrm{joint\_deviation\_legs}}$
& $\|\mathbf{q}_{\mathrm{legs}}-\mathbf{q}^{\mathrm{def}}_{\mathrm{legs}}\|_1$
& -1.0 \\

$\mathcal{R}_{\mathrm{flat\_orientation}}$
& $\|\mathbf{g}^{b}_{xy}\|_2^2$
& -5.0 \\

$\mathcal{R}_{\mathrm{base\_height}}$
& $(h-h^\star)^2,\quad h^\star=0.78$
& -10.0 \\

$\mathcal{R}_{\mathrm{gait}}$
& $\sum_i \mathbbm{1}\!\left[c_i=\mathbbm{1}(\phi_i<0.55)\right]\mathbbm{1}\!\left[\|\mathbf{v}^{\mathrm{cmd}}\|_2>0.1\right],\quad \phi_i=(\phi+\Delta_i)\bmod 1$
& 0.5 \\

$\mathcal{R}_{\mathrm{feet\_slide}}$
& $\sum_i \|\mathbf{v}^{\mathrm{foot},xy}_i\|_2\,\mathbbm{1}[c_i=1]$
& -0.2 \\

$\mathcal{R}_{\mathrm{feet\_clearance}}$
& $\exp\!\left(-\frac{\sum_i (z_i-z^\star)^2 \tanh(2\|\mathbf{v}^{\mathrm{foot},xy}_i\|_2)}{0.05}\right),\quad z^\star=0.1$
& 1.0 \\

$\mathcal{R}_{\mathrm{undesired\_contacts}}$
& $\sum_{j\in\mathcal{B}_{\mathrm{non\text{-}ankle}}}\mathbbm{1}[F_j>1]$
& -1.0 \\

\bottomrule
\end{tabular}
}
\end{table*}

\subsection{Training Hyperparameters}
\label{app:additional_experiments}
Table~\ref{tab:agent_hparams} reports the hyperparameters used for PPO, GenPO, and SteerGenPO. PPO, GenPO, and SteerGenPO Stage 1 use the same rollout and optimization settings whenever applicable, so that the comparison mainly reflects the difference between the standard actor-critic policy and the flow-based policy parameterization.

SteerGenPO Stage 2 further fine-tunes the Stage-1 policy in the latent action space. Because latent-space optimization is more sensitive to aggressive updates, we use a more conservative fine-tuning setup with a smaller learning rate, reduced exploration noise, and latent-specific regularization. All numerical settings are provided in Table~\ref{tab:agent_hparams}.

\begin{figure}[ht]
    \centering
    \includegraphics[width=0.5\linewidth]{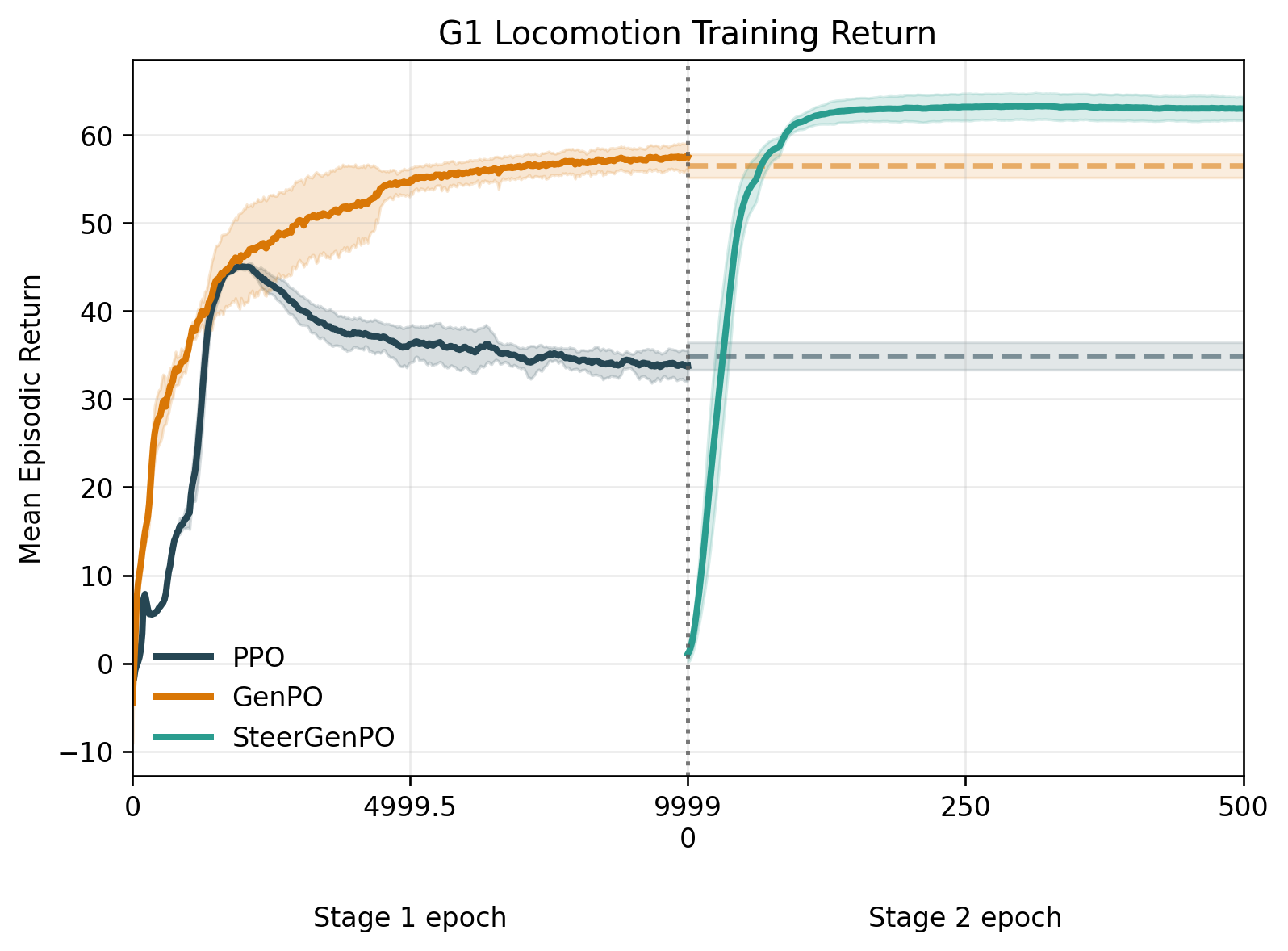}
    \caption{Training curves on Unitree-G1 locomotion task. Stage 1 presents the training curves of both PPO and GenPO. Stage 2 shows the second-stage curve of SteerGenPO.}
    \label{fig:g1_reward}
\end{figure}

Figure~\ref{fig:g1_reward} illustrates the corresponding training process. Stage 1 shows that both PPO and GenPO are trained from scratch, and Stage 2 demonstrates that SteerGenPO, initialized from the converged GenPO policy of Stage 1, continues to improve the reward.

\begin{table*}[ht]
\centering
\caption{Agent and training hyperparameters for PPO, GenPO, and SteerGenPO. SteerGenPO is trained in two stages: Stage 1 adopts the same configuration as GenPO, while Stage 2 performs latent fine-tuning with a separate set of hyperparameters.}
\label{tab:agent_hparams}
\resizebox{\textwidth}{!}{
\begin{tabular}{lcccc}
\toprule
& & & \multicolumn{2}{c}{SteerGenPO} \\
\cmidrule(lr){4-5}
Hyperparameter & PPO & GenPO & Stage 1 & Stage 2 \\
\midrule
Max iterations & 20000 & 20000 & 20000 & 10000 \\
Steps per env & 24 & 24 & 24 & 24 \\
Learning epochs & 5 & 5 & 5 & 5 \\
Mini-batches & 4 & 4 & 4 & 4 \\
Learning rate & 1.0e-3 & 1.0e-3 & 1.0e-3 & 2.0e-4 \\
Schedule & adaptive & adaptive & adaptive & adaptive \\
$\gamma$ & 0.99 & 0.99 & 0.99 & 0.99 \\
$\lambda$ & 0.95 & 0.95 & 0.95 & 0.95 \\
Entropy coefficient & 0.01 & 0.01 & 0.01 & 0.003 \\
Desired KL & 0.01 & 0.01 & 0.01 & 0.01 \\
Clip parameter & 0.2 & 0.2 & 0.2 & 0.2 \\
Value loss coefficient & 1.0 & 1.0 & 1.0 & 1.0 \\
Max grad norm & 1.0 & 1.0 & 1.0 & 1.0 \\
Initial action noise std & 1.0 & 1.0 & 1.0 & 0.2 \\
Actor hidden dims & [512, 256, 128] & [512, 256, 128] & [512, 256, 128] & [1024, 512, 256] \\
Critic hidden dims & [512, 256, 128] & [512, 256, 128] & [512, 256, 128] & [1024, 512, 256] \\
Activation & ELU & ELU & ELU & ELU \\
\midrule
Flow steps & -- & 5 & 5 & -- \\
Flow mixing parameter & -- & 0.95 & 0.95 & -- \\
Time embedding dim & -- & 32 & 32 & -- \\
Time MLP dims & -- & [64, 64] & [64, 64] & -- \\
Flow std & -- & 3.0 & 3.0 & -- \\
\midrule
Latent penalty scale & -- & -- & -- & 0.03 \\
Latent initstd & -- & -- & -- & 0.1 \\
\bottomrule
\end{tabular}
}
\end{table*}

\end{document}